\documentclass[letterpaper, 10 pt, conference]{ieeeconf}  

\IEEEoverridecommandlockouts                              
\usepackage{graphicx}
\usepackage{verbatim}
\usepackage{array}
\usepackage{upgreek}
\usepackage{float}   
\usepackage{amsmath,amssymb}
\usepackage{cite}
            
\usepackage{booktabs}
\usepackage{algorithm}
\usepackage[noend]{algpseudocode}

\usepackage{todonotes}
\usepackage{xcolor}

\usepackage[normalem]{ulem}
\let\labelindent\relax
\usepackage{enumitem}

\makeatletter
\def\algbackskip{\hskip-\ALG@thistlm}
\makeatother

\usepackage{tikz}
\usetikzlibrary{positioning, shapes, arrows.meta}
\usetikzlibrary{shapes,arrows,chains}
\usetikzlibrary{arrows,calc,automata}

\tikzset{
    block/.style = {draw, rectangle, 
        minimum height=.5cm, 
        minimum width=1cm},
    input/.style = {coordinate,node distance=.5cm},
    output/.style = {coordinate,node distance=.5cm},
    arrow/.style={draw, -latex,node distance=.5cm},
    pinstyle/.style = {pin edge={latex-, black,node distance=1cm}},
    sum/.style = {draw, circle, node distance=1cm},
    line/.style={-{Stealth}}
    }

\title{\LARGE \bf
Hoverflie: An empirical investigation of rotor shrouds to transform micro air vehicles into multi-modal hovercraft
}

\author{
Mrinmoy Modak and Daniel S. Drew 
\thanks{Department of Electrical and Computer Engineering, University of Hawaii at Manoa, Honolulu, HI 96822, USA}
\thanks{Corresponding author: Daniel S. Drew, \tt{ddrew@hawaii.edu}}
}

\begin{document}

\maketitle
\thispagestyle{empty}
\pagestyle{empty}

\begin{abstract}
Small rotorcraft intended for use indoors or around the built environment have extremely limited flight duration. This paper presents the design and experimental characterization of a custom shroud system that transforms a Crazyflie 2.1 micro air vehicle into a multi-modal robot capable of operating as a high-efficiency hovercraft or a free-flying drone. 
A custom experimental platform was developed for precise control of hover height and rotor duty cycle, and automated data logging of lift forces. 
Parametric testing of duct, intake, and nozzle geometries was performed to investigate the impact of shroud configuration on in-ground-effect and free-flight performance. 
An empirical model is developed which, unlike typical models for ground effect in rotorcraft, captures the suckdown effect that reduces force at intermediate height.
It is shown that, through proper design of the shroud, beneficial ground effects can be increased while diminishing negative effects both close to the ground and in free-flight. 
An optimized configuration exhibited nearly three times higher in-ground-effect force while maintaining comparable out-of-ground-effect aerodynamic thrust, although the added shroud mass reduces free-flight control authority.
Lightweight shrouds are manufactured using thin-film thermoformed components, and total single-charge flight time is shown to increase by 60\% in-ground-effect while decreasing by only 30\% in free-flight as compared to the stock drone. Finally, controlled flight in the air, hovering close to the ground, and hover-to-flight transitions are demonstrated using a simple mode-switching controller, with tracking errors reported to quantify performance.
This work provides an experimentally-validated and easily adoptable foundation for future research into lightweight ground-effect vehicles and hybrid drone-hovercraft systems.
\end{abstract}

\section{Introduction}
\label{sec:introduction}
Micro air vehicles (MAVs) have emerged as a compelling class of aerial robot due to their small size, low cost, and ability to operate in confined, cluttered environments~\cite{floreano_science_2015}. They are envisioned for applications including indoor surveillance, search and rescue, infrastructure inspection, and warehousing~\cite{kumar_opportunities_2012}. However, MAVs operate in a low Reynolds number regime where aerodynamic efficiency is inherently limited and induced losses are significant. Combined with limited onboard energy storage, these factors severely constrain endurance and payload, making efficiency improvement a central challenge in MAV design~\cite{mulgaonkar_power_2014}. 

One promising but underexplored opportunity lies in proximity-induced aerodynamic effects. When a rotorcraft operates near a surface, interactions with the ground modify the surrounding flow field and can significantly augment lift while reducing power consumption~\cite{matus-vargas_ground_2021, 21}. This ground effect is the operating principle of hovercraft, which use pressurized air cushions to achieve efficient near-surface locomotion. While the ground effect has been extensively studied for full-scale helicopters \cite{1}, quadrotors~\cite{2, 3, 4, 5}, and traditional bag-skirt hovercraft~\cite{6, 7, 8}, there has been no systematic empirical investigation of MAV-scale multirotors capable of both surface-assisted and free-flight operation.

\begin{figure}[t]
    \centering
    \includegraphics[width=1\linewidth]{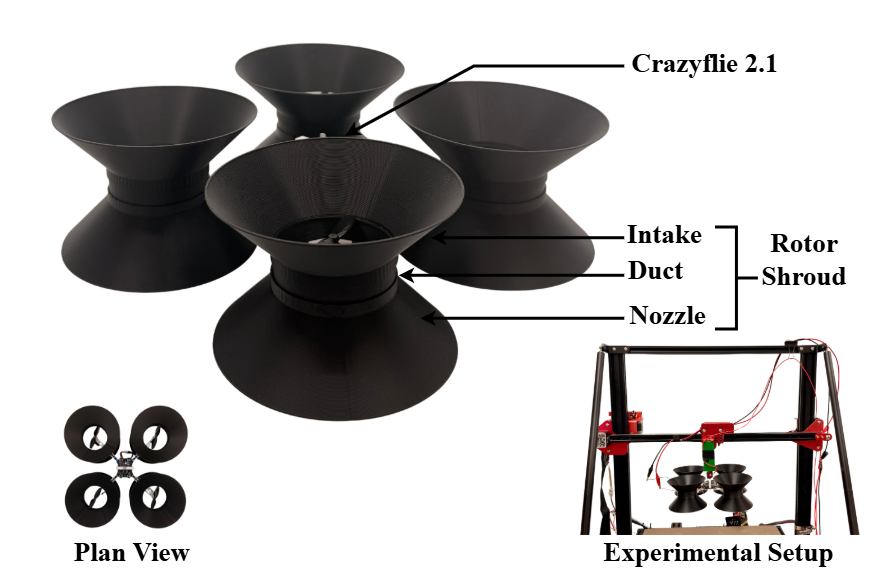}%
    \caption{The Hoverflie, a Crazyflie-based multi-modal hovercraft, uses lightweight rotor shrouds which are designed to maximize in-ground-effect thrust while minimizing impact on free-flight performance. The inset shows the experimental setup for thrust measurement at different heights.}
    \label{fig:hovercraft}
    \vspace{-2em}
\end{figure}

Versatile, well-supported research platforms such as the Crazyflie 2.1 make it possible to experimentally explore ground effect aerodynamics at the MAV scale. However, integrating shrouds into multirotor MAVs introduces aerodynamic complications, including flow recirculation, back pressure, separation, and the suckdown effect, which can negate proximity-induced lift benefits and degrade out-of-ground-effect performance~\cite{han_aerodynamic_2019}. These adverse interactions are especially critical during transitions between near-ground hovering and free-flight, where hybrid mobility would otherwise offer substantial endurance advantages.

In this work, we investigate whether a lightweight rotor-integrated shroud system can transform a quadrotor MAV into a multi-modal hovercraft while preserving acceptable free-flight capability. We convert a Crazyflie 2.1 into ``Hoverflie'' (Fig.~\ref{fig:hovercraft}) and develop a vertical test platform using a load cell integrated with a programmable gantry to systematically measure lift as a function of normalized height and shroud geometry. Through parametric variation of duct, nozzle, and intake configurations, we quantify how geometry influences in-ground-effect lift augmentation, out-of-ground-effect performance, and suckdown forces at transition heights.

Our results show that appropriate shroud design can increase in-ground-effect lift by nearly three times while maintaining comparable out-of-ground-effect thrust. However, the additional mass introduced by the shroud reduces the thrust-to-weight ratio and available control authority, particularly in yaw, relative to the stock Crazyflie. An optimized configuration achieves approximately 60$\%$ endurance improvement in-ground-effect with a 30$\%$ reduction in free-flight endurance. We further demonstrate stable hover-to-flight transitions and trajectory tracking, establishing the feasibility of hybrid ground-effect–aerial locomotion.

The primary contributions of this work are: 1) A reproducible experimental methodology for high-resolution characterization of ground effect forces in a multirotor MAV; 2) A systematic empirical mapping between shroud geometry (duct, nozzle, intake) and aerodynamic performance; 3) Development of an empirical model that captures the suckdown effect in shrouded multirotors, enabling quantitative design comparisons; 4) Demonstration of a lightweight multi-modal hovercraft capable of trajectory tracking, including hover-to-flight transitions; and 5) Quantified endurance trade-offs showing substantial energy savings in-ground-effect with a penalty in free-flight. Together, these findings provide an experimentally-validated foundation for hybrid hovercraft–drone systems capable of energy-efficient near-surface operation while retaining full aerial mobility when required.

\section{Related Work}
\label{sec:Related Work}

\subsection{The Ground Effect in Small-scale Rotorcraft}
The majority of ground effect models in use today are based on closed-form relationships between thrust augmentation and rotor height that were derived using traditional analytical methods based on potential flow and image theory~\cite{9}. Subsequent experimental tests confirmed these findings, revealing that the ground effect is most significant when the rotor height is between one and two rotor radii, resulting in lower power consumption during hover and landing~\cite{1}. However, these early studies were primarily concerned with full-scale single-rotor helicopters and did not consider small-scale or multi-rotor effects. Recent work has demonstrated that classical helicopter models underestimate the ground effect in multirotor platforms; experimental results using quadrotor test benches show that thrust augmentation in multirotors can persist up to five rotor radii, significantly exceeding single-rotor predictions~\cite{3}. Additional studies on small UAV rotors indicate that low Reynolds number effects and rotor geometry strongly influence ground effect magnitude, limiting the direct applicability of large-scale empirical models~\cite{10}.

Ground and other proximity effect behavior has also been investigated for ducted rotor systems. Studies have demonstrated that as ground clearance decreases, the thrust contribution from the duct diminishes while rotor and stator blade thrust increases, yielding a net total thrust gain~\cite{32}. Prior work on coaxial ducted-fan aerial robots has shown that conventional ground effect models are not directly suitable for ducted-fan aircraft, and developed new experimentally-fit models to describe their performance~\cite{29}. Other studies used experimentally validated three-dimensional numerical models to show that ground effect becomes significant when the system altitude is below one duct length~\cite{30}, and that ground blockage can reduce mass flow rate, duct thrust, and overall aerodynamic performance. In addition, ducted-fan operation in combined ground–wall corner environments has been modeled using artificial neural networks, showing that proximity effects can generate nonlinear aerodynamic forces and moments that are difficult to capture using simple analytical fitting methods~\cite{31}. In contrast with these prior studies, we specifically investigate multi-part shrouds comprising intakes, ducts, and nozzles, which have effects which are distinct from purely ducted fans (e.g., nozzle-induced suckdown forces). 

\subsection{Proximity Effects in Surface-Assisted Flight}
Recent studies on proximity effect-mediated flight demonstrate that aerodynamic interactions with nearby boundaries can be exploited for stability enhancement, energy reduction, hybrid aerial–surface mobility, contact-enabled behaviors, and proximity-aware motion planning~\cite{22, 24, 25, 26}. 
Wall and ceiling proximity effects have been leveraged for passive lateral stabilization, perching, and extended operation, while empirical and parametric analyses of ground, wall, and ceiling interactions have quantified force augmentation and stability variations near surfaces~\cite{21, 27, 28}. In contrast with these examples, we explore structural modifications to an existing platform which can improve proximity effect performance while minimizing free-flight impact. 

\subsection{Hovercraft Design}
Early work on hovercraft platforms concentrated on system design and prototyping, stressing practical issues with lift generation, bag and finger shroud design, and propulsion integration~\cite{6, 11}. To increase operational efficiency, optimization-based approaches have been used, for example resulting in performance gains via evolutionary algorithms~\cite{7}. Robust and nonlinear control solutions have been devised to overcome the modeling uncertainties and external disruptions inherent in hovercraft dynamics \cite{8,12}. Small-scale autonomous hovercraft have also been created, typically as platforms to study new control paradigms~\cite{roubieu_fully-autonomous_2012}. 
\begin{figure} [b]
    \centering
    \colorbox{gray!10}{\includegraphics[width=0.96\linewidth]{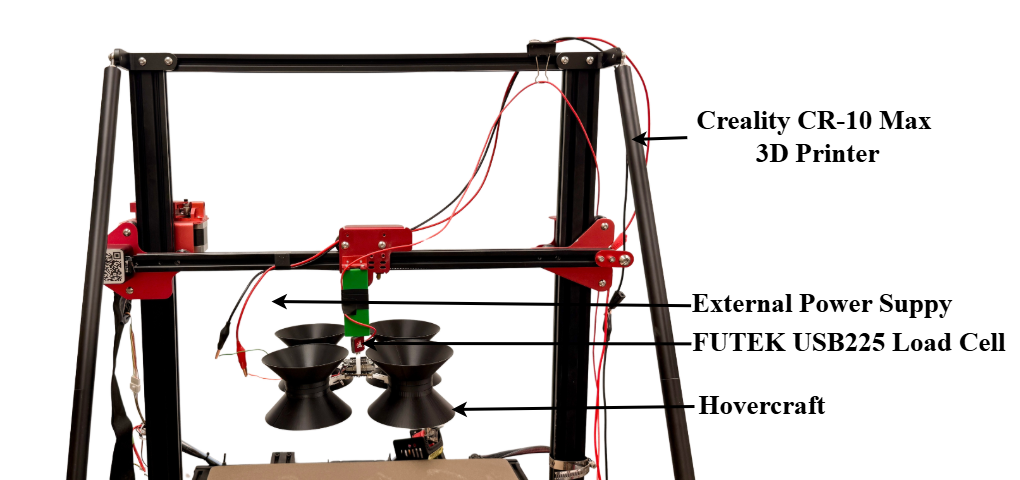}}
    \caption{Integrated test setup for systematic thrust measurement at different distances from the ground plane.}
    \label{fig:setup}
    \vspace{-1em}
\end{figure}

In contrast to previous hovercraft research, we concentrate on duct-nozzle-intake-based hovercraft geometries rather than traditional bag or finger skirt designs. While we investigate the best geometric configurations to increase hover efficiency and stability near the ground, we also look toward hybrid hover-and-fly capability on a single platform. This necessitates exploring how shroud geometry can minimize the suckdown effect and other near-ground aerodynamic interactions, which are largely ignored in existing work.

\section{Methods}
\label{sec:Methods}

\subsection{Experimental Setup and System Integration}

An experimental platform was constructed to characterize the ground effect forces produced by the vehicle. The system integrates a Crazyflie 2.1, a FUTEK LSB201 S-Beam load cell, and a Creality CR-10 Max 3D printer. The load cell was rigidly attached to the printer’s gantry, and the vehicle was equipped with a 3D-printed attachment which fits onto the central fuselage and can be screwed into the load cell, as shown in Fig.~\ref{fig:setup}. The printer's z-axis stepper motor enables exact vertical positioning of the vehicle relative to the ground plane. A constant external power supply provided stable input to the Crazyflie during force measurements. A custom Python control panel with a graphical interface coordinated all subsystems in real time. The interface allows users to: set and log output data, command the Crazyflie’s thrust level through a PWM signal (0–65535 range), and move the printer gantry in different height steps with fixed intervals.

\subsection{Shroud Geometry}

\begin{figure} [t]
    \centering
    \colorbox{gray!0}{\includegraphics[width=0.9\linewidth, trim= 0cm 0cm 0cm 0cm, clip]{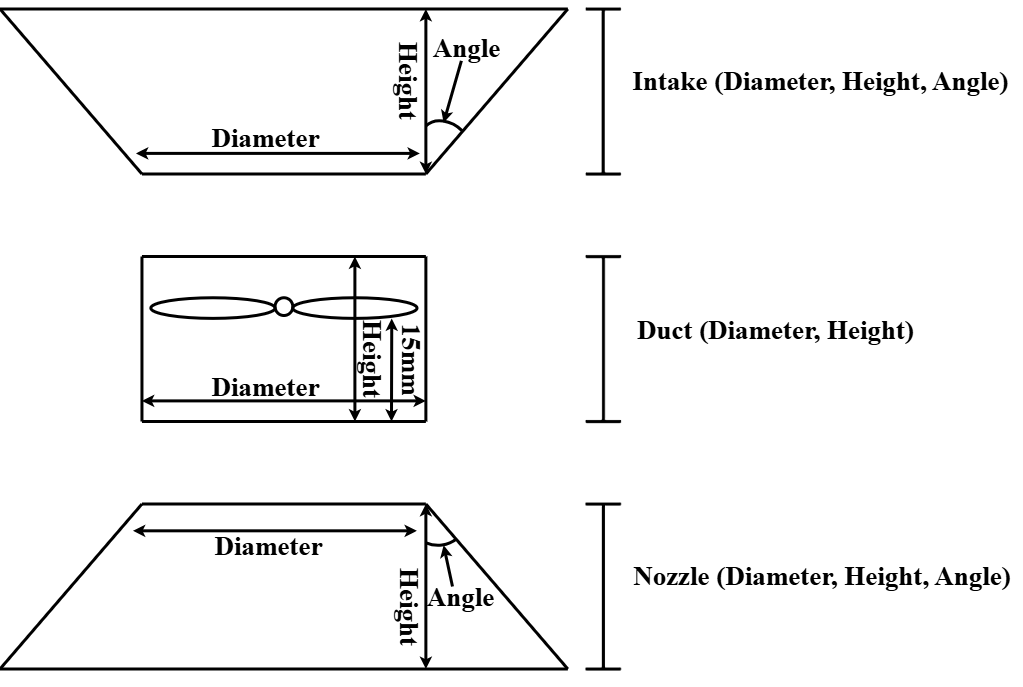}}
    \caption{Geometric parameters of the shroud components, including diameter, height, and angle for the duct, intake, and nozzle sections.}
    \label{fig:geometry}
    \vspace{-1em}
\end{figure}

The lift characteristics of the hovercraft were found to be highly sensitive to the geometry of the shroud surrounding the propeller. To systematically investigate this effect, the shroud was divided into three functional regions - duct (D), nozzle (N), and intake (I), each parameterized by its diameter, height, and angular profile, as shown in Fig.~\ref{fig:geometry}. For brevity, the geometric notation used throughout this section follows the form D (Diameter, Height) + N (Diameter, Height, Angle) + I (Diameter, Height, Angle).

\begin{figure}
    \centering
    \includegraphics[width=1\linewidth, trim= 0cm 0.8cm 0cm 0.2cm, clip]{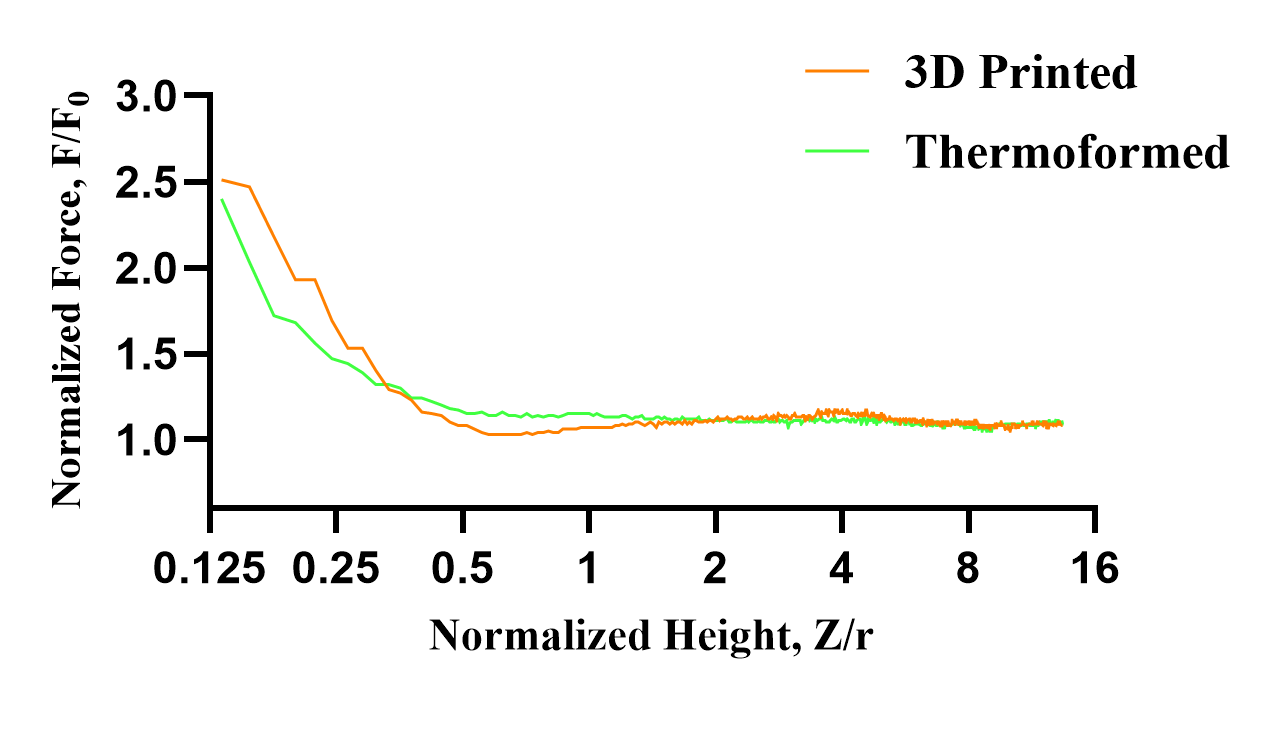}
    \caption{Normalized thrust ratio versus normalized height illustrating the aerodynamic trendline of the finalized shroud geometry D(47,15) + N(47,15,45) + I(47,15,45) fabricated using 3D printing and thermoforming.}
    \label{3D vs Thermoformed}
    \vspace{-1em}
\end{figure}

\begin{figure}
    \centering
    \colorbox{gray!0}{\includegraphics[width=1\linewidth]{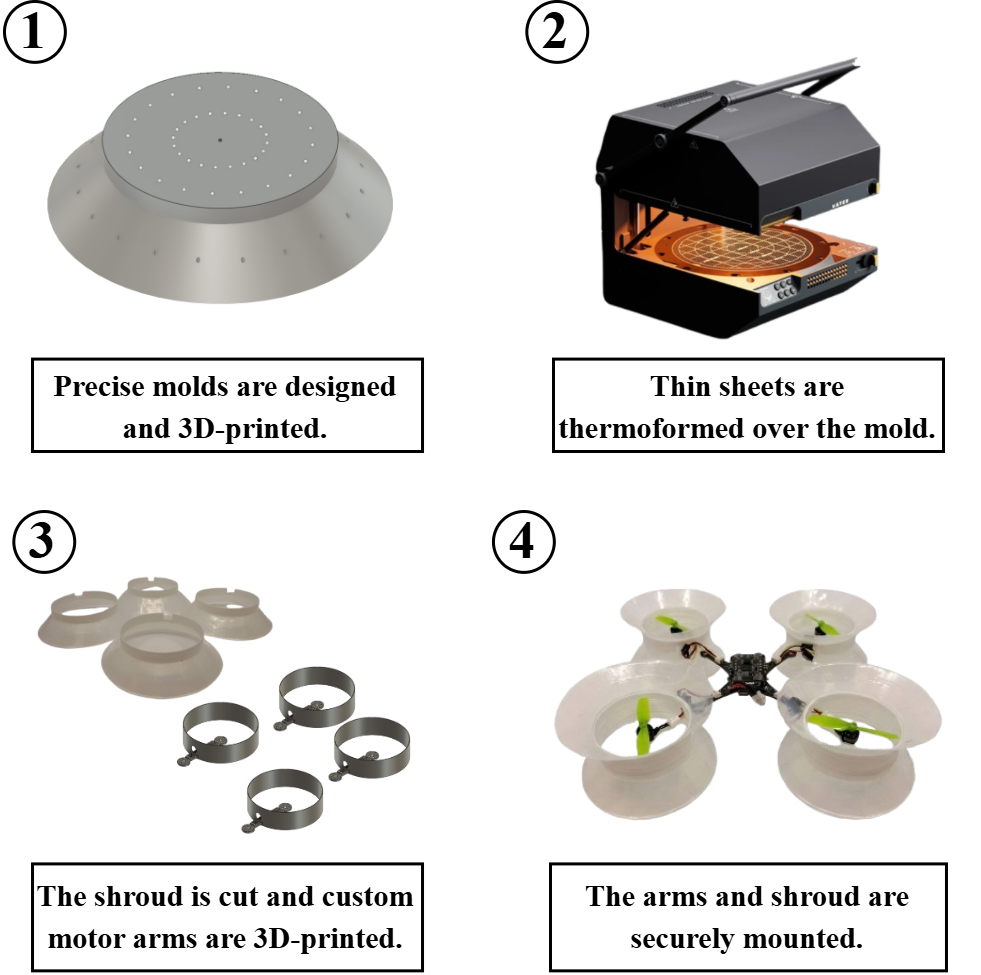}}
    \caption{The hovercraft shrouds are fabricated with a combination of 3D printing and thermoforming and can be quickly fixed to the Crazyflie chassis.}
    \label{fig:fabrication}
    \vspace{-1.5em}
\end{figure}

\begin{figure*}
    \centering
    \includegraphics[width=1\linewidth, trim= 0.5cm 0cm 0.5cm 0.1cm, clip]{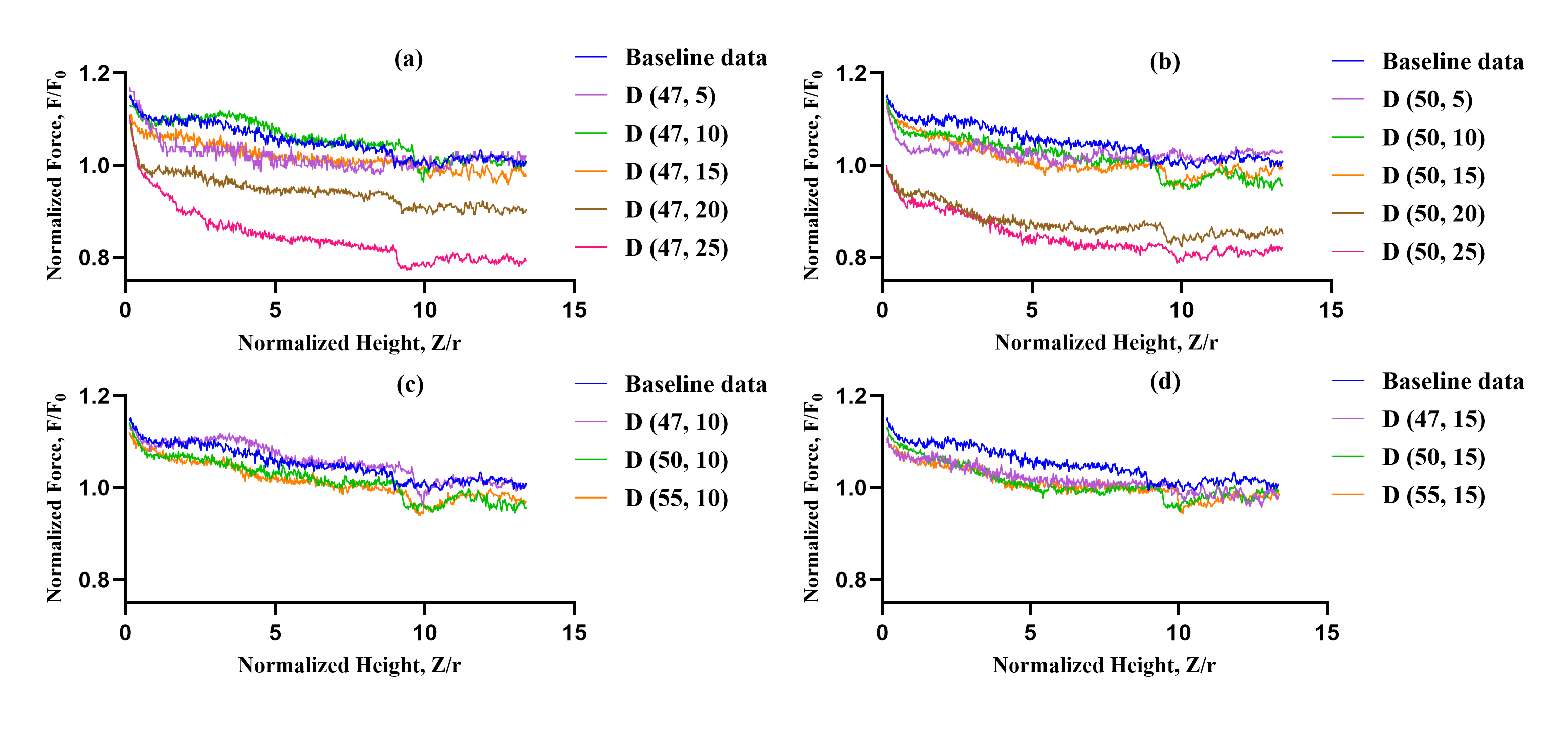}
    \vspace{-4em}
    \caption{Normalized thrust ratio versus normalized height illustrating the influence of duct geometry on ground effect performance relative to the baseline data. (a) Variation of duct height at 47 mm diameter. (b) Variation of duct height at 50 mm diameter. (c) Variation of duct diameter at 10 mm duct height. (d) Variation of duct diameter at 15 mm duct height.}
    \label{fig:duct_geometry}
    \vspace{-1em}
\end{figure*}

\subsection{Fabrication Process}

The shrouds used in geometric variation experiments were fabricated using a 3D printer to ensure rapid and easy prototyping. After the geometry was finalized, the shrouds were thermoformed to reduce structural weight while maintaining similar aerodynamic performance (Fig.~\ref{3D vs Thermoformed}). While a hypothesis is that the remaining difference is a result of variation in surface roughness and edge quality between the two fabrication methods, further investigation is needed.

To thermoform the custom shroud for the Crazyflie, positive molds were first created using a 3D printer to capture the desired geometrical shape. Thin polymer films (e.g., 0.005'' ABS) were then shaped over these molds using a Mayku Multiplier pressurized thermoforming machine, allowing precise replication of the individual shroud sections. After forming, the parts were carefully singulated and trimmed using scissors. The final shroud was then assembled and securely mounted onto the Crazyflie using custom motor arms with mechanical affordances for attachment, ensuring consistent alignment and fit during flight tests. Fig.~\ref{fig:fabrication} illustrates the fabrication and assembly process.

All relevant design files, tutorials for reproduction of the Hoverflie
platform, and further instructions on reproducing the experiments are available
at LINKS REMOVED FOR DOUBLE-BLIND REVIEW or upon request.

\section{Results and Analysis}
\label{sec:Results and Analysis}
\subsection{Shroud Design and Geometric Optimization}

Shroud optimization was performed in a staged manner. We first varied the duct diameter and height to identify configurations that maximize lift augmentation in-ground-effect. A nozzle and intake were then introduced, and their geometric parameters were refined to improve thrust continuity across the in-ground-effect (IGE) to out-of-ground-effect (OGE) transition region. All configurations were evaluated relative to the stock (i.e., without structural modifications) Crazyflie 2.1 to ensure consistent comparison.

\textit{Duct Diameter and Height:} To first determine the optimal duct height, experiments were conducted using duct diameters of 47 mm and 50 mm. As shown in Fig.~\ref{fig:duct_geometry}(a) and Fig.~\ref{fig:duct_geometry}(b), duct heights of 5 mm, 10 mm and 15 mm produced the highest normalized lift across the tested height range. Since a minimum duct height of 10 mm was required for mechanically secure integration of the nozzle and intake, subsequent experiments focused on the 10 mm and 15 mm duct configurations. The diameter was then varied among 47 mm, 50 mm, and 55 mm to assess its influence on thrust. As shown in Fig.~\ref{fig:duct_geometry}(c) and Fig.~\ref{fig:duct_geometry}(d), the 47 mm diameter consistently produced the highest thrust and was therefore selected for subsequent testing. This configuration provides the tightest clearance (the nominal Crazyflie propeller diameter is 45~mm), which in theory reduces blade-tip vortex formation, a dominant source of losses in small rotorcraft.

\begin{figure*}
    \centering
    \includegraphics[width=1\linewidth, trim= 0.5cm 0cm 0.5cm 0.6cm, clip]{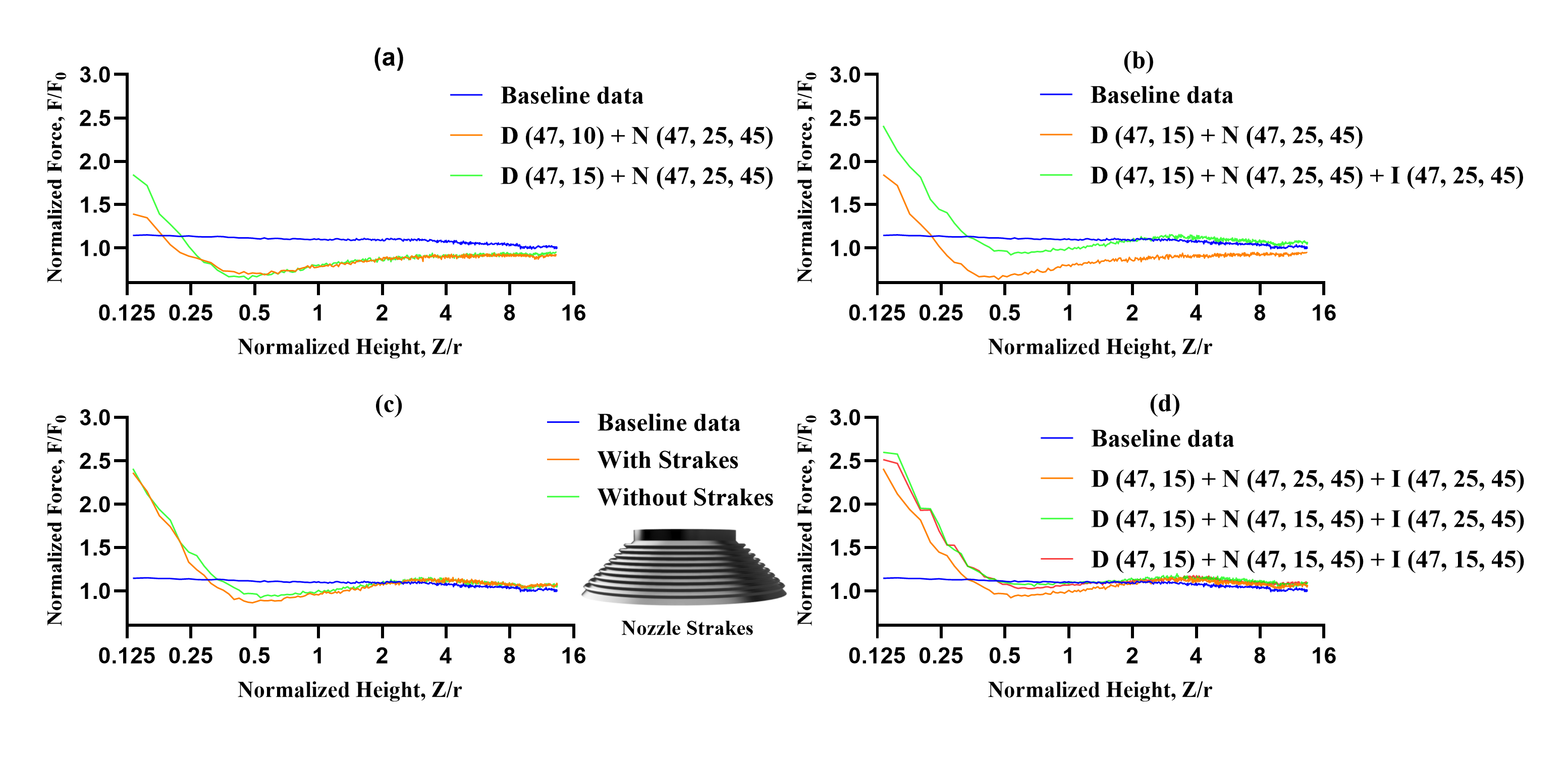}
    \vspace{-4em}
    \caption{Normalized thrust ratio versus normalized height illustrating the influence of shroud configuration on ground-effect performance relative to the baseline data. (a) Added nozzle with variation of duct height. (b) Added intake. (c) Comparison with and without strakes (Tested Geometry: D (47,15) + N (47,25,45) + I (47,25,45)) (d) Variation of both nozzle and intake heights.}
    \label{fig:suckdown_mitigation}
    \vspace{-1em}
\end{figure*}

\begin{figure} [H]
    \centering
    \includegraphics[width=0.9\linewidth]{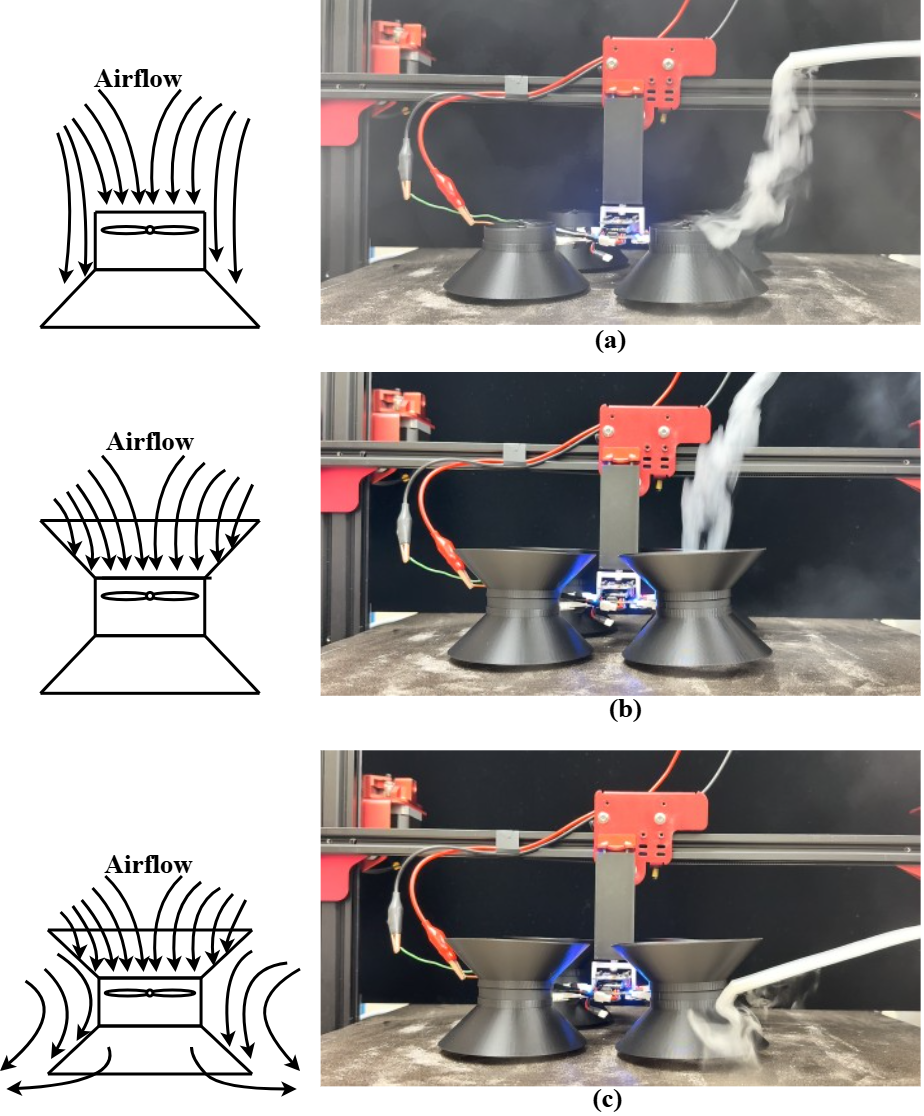}
    \vspace{-1em}
    \caption{Illustrations and smoke tests to investigate aerodynamic mechanisms of observed suckdown. (a) Air bypassing the duct. (b) Air successfully captured by an included intake. (c) Ambient air entrained by the wall jet to flow along the nozzle surface.}
    \label{fig:smoketest}
    \vspace{-1em}
\end{figure}

\textit{Suckdown Mitigation:} A nozzle was added to enhance IGE thrust. Initial tests were performed using a 47 mm duct with heights of 10 mm and 15 mm (Fig.~\ref{fig:suckdown_mitigation}(a)). The configuration in which the duct lip was level with the propeller plane (15 mm) yielded the strongest $F_{IGE}$, but exhibited reduced $F_{OGE}$ performance. Smoke visualization (Fig.~\ref{fig:smoketest}(a)) revealed that part of the incoming airstream escaped the duct and was entrained to flow along the nozzle surface, resulting in a ``suckdown'' effect producing a dip in thrust which was dominant at transition heights between in-ground-effect and out-of-ground-effect operation.

To mitigate this effect and recover low-altitude free-flight performance, an intake section was added above the duct to redirect incoming flow and reduce external entrainment (Fig.~\ref{fig:smoketest}(b)). As shown in Fig.~\ref{fig:suckdown_mitigation}(b), the intake significantly improved $F_{OGE}$ and reduced suckdown magnitude.

Although the introduction of an intake eliminated the problem of incoming air being diverted around the duct, the impinged wall jet generated by the exhaust still entrains ambient air to flow along the nozzle surface (Fig.~\ref{fig:smoketest}(c)). Strakes were introduced along the nozzle in an attempt to prevent a boundary layer of this entrained air from attaching to the nozzle. Fig.~\ref{fig:suckdown_mitigation}(c) shows that this modification produced a negligible difference in thrust, indicating an attached boundary layer on the nozzle is unlikely to be the dominant remaining source of suckdown. With the hypothesis that suckdown instead originates from pressure loading on the outer nozzle surface, the nozzle surface area was then reduced. This geometric modification substantially decreased the magnitude of the suckdown dip and improved thrust consistency across transition heights, as shown in Fig.~\ref{fig:suckdown_mitigation}(d), while intake height was shown to have negligible effect.

\begin{figure}
    \centering
    \includegraphics[width=1\linewidth, trim= 0.5cm 0cm 0.5cm 0.6cm, clip]{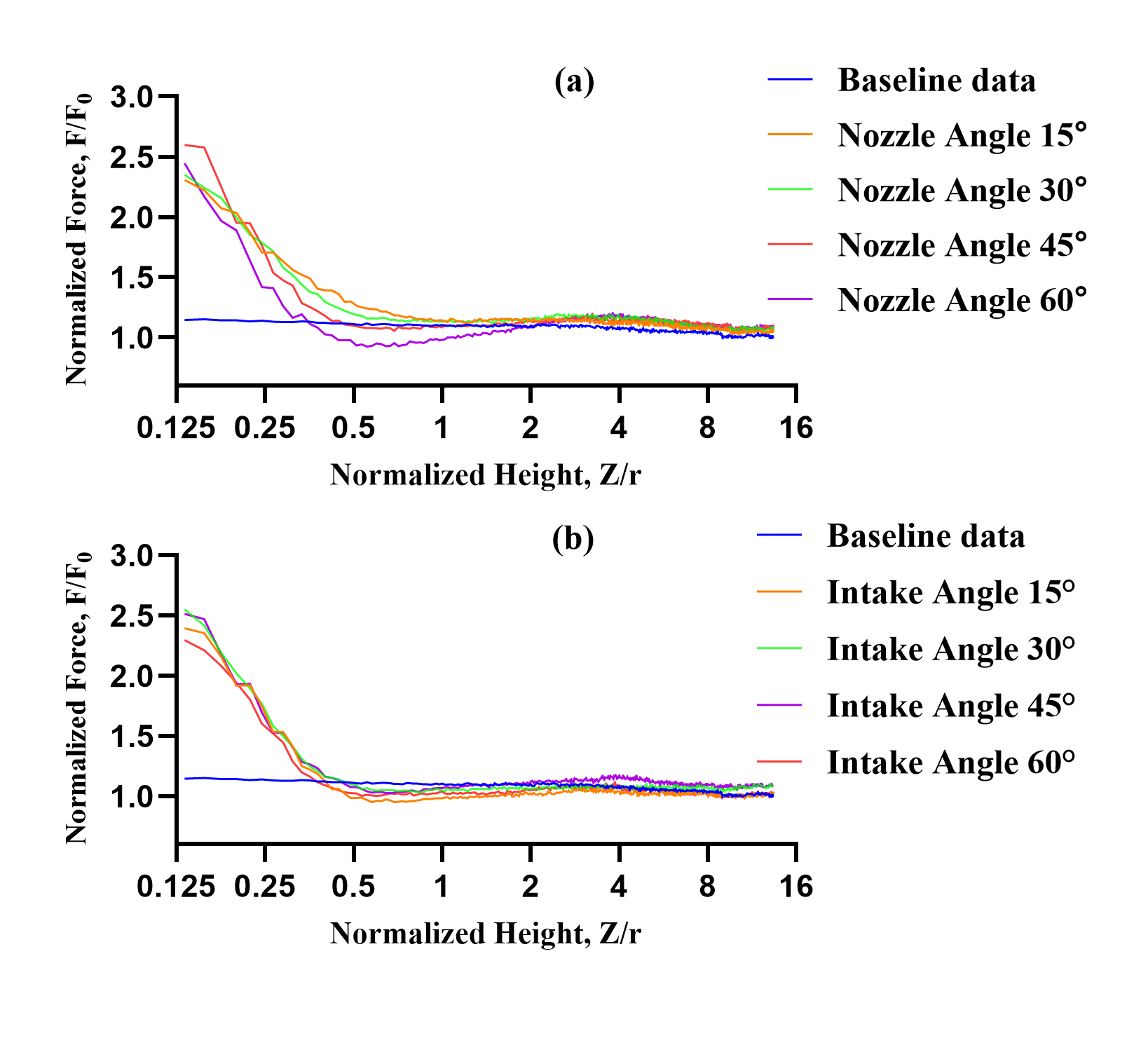}
    \vspace{-4em}
    \caption{Normalized thrust ratio versus normalized height as a function of nozzle and intake angles. (a) Variation of nozzle angle with constant duct and intake configuration. (Tested Geometry: D (47,15) + N (47,25, $\theta)$ + I (47,25,45)) (b) Variation of intake angle with fixed duct and nozzle configuration (Tested Geometry: D (47,15) + N (47,15,45) + I (47,15, $\theta$)).}
    \label{fig:angles}
    \vspace{-2em}
\end{figure}

\textit{Nozzle and Intake Angle:} The influence of angular parameters was then investigated. Increasing the nozzle angle produced a clear enhancement in $F_{IGE}$, as shown in Fig.~\ref{fig:angles}(a). A larger nozzle angle expands the effective air-cushion area beneath the vehicle, increasing pressure buildup and lift augmentation near the ground. However, this benefit is accompanied by a corresponding increase in suckdown force during transition to OGE operation. 

In contrast, variation of the intake angle produced only minor changes in both $F_{IGE}$ and $F_{OGE}$ (Fig.~\ref{fig:angles}(b)), indicating that intake geometry plays a secondary aerodynamic role compared to nozzle angle. Consequently, a minimal-mass intake configuration was selected to prioritize overall weight reduction without sacrificing performance.

\begin{figure}
    \centering
    \includegraphics[width=1\linewidth]{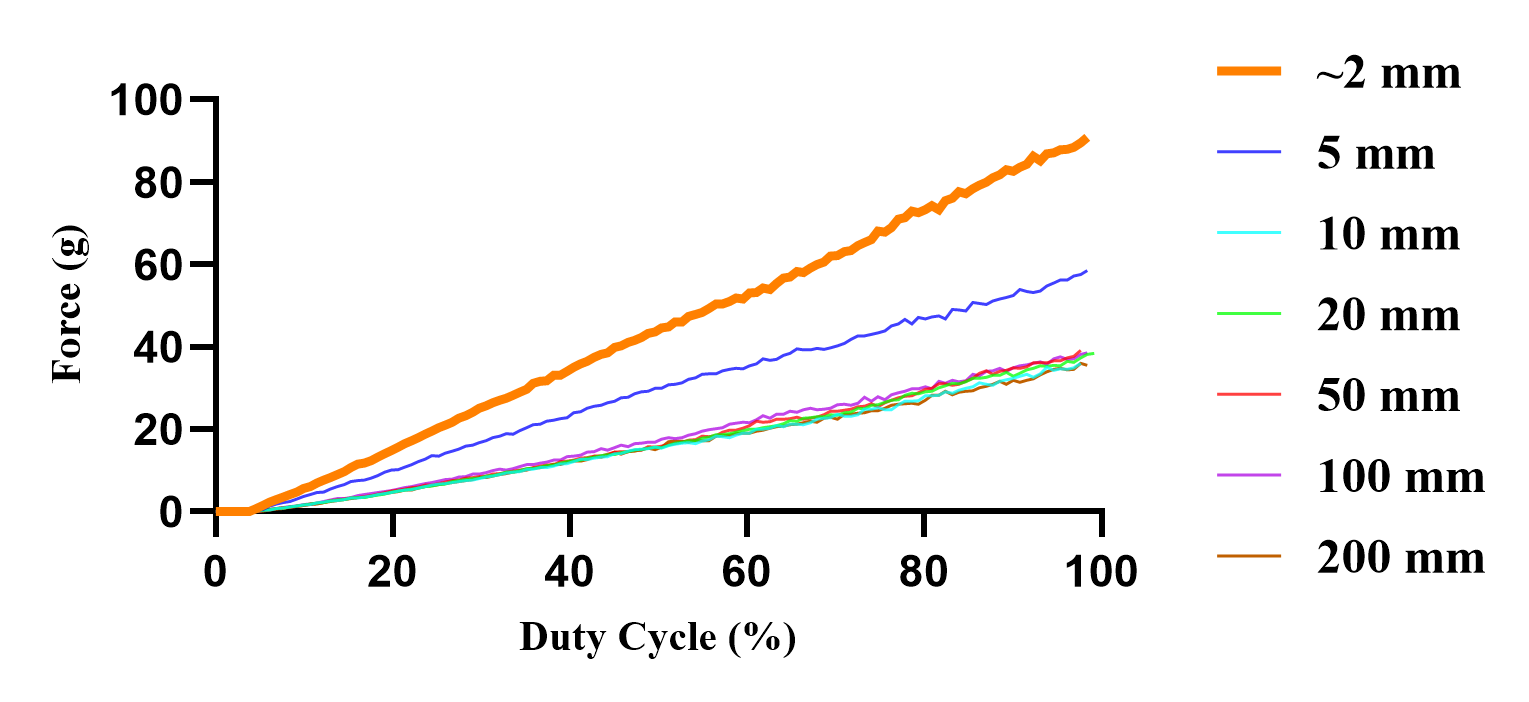}
    \vspace{-2em}
    \caption{Measured thrust force as a function of PWM duty cycle at different rotor-ground clearances for the selected shroud design used in subsequent control and endurance validation.}
    \label{fig:pwm}
    \vspace{-1.5em}
\end{figure}

\textit{Selected Configuration Characterization:} Balancing near-ground lift enhancement against transition stability, the final optimized shroud configuration was selected as \text{D}(47,15) + \text{N}(47,15,45) + \text{I}(47,15,45). To further characterize thrust authority across operating regimes for the selected shroud geometry, we measured thrust as a function of motor PWM duty cycle at multiple rotor-ground clearances (Fig.~\ref{fig:pwm}). As height increases, the slope of the thrust–duty cycle relationship decreases and approaches a constant value, reflecting the diminishing proximity-induced lift. These results confirm that the optimized configuration provides sufficient thrust margin for hover-to-flight transitions while benefiting from substantial lift amplification, and therefore payload potential, in near-ground operation.

\subsection{Model Validation and Curve Fitting}
\begin{figure}
    \centering
    \includegraphics[width=1\linewidth]{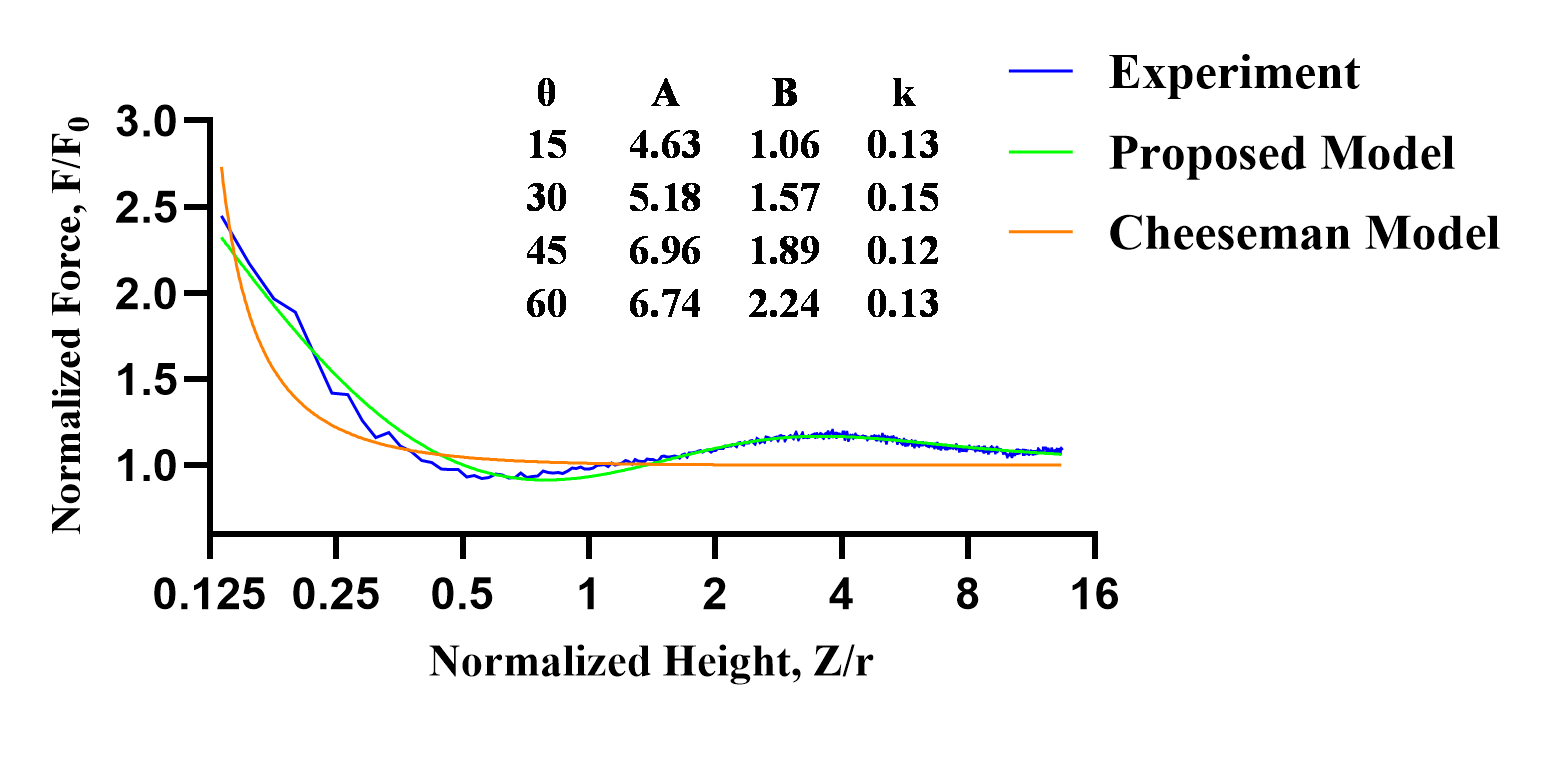}
    \vspace{-3em}
    \caption{Normalized thrust ratio vursus normalized height, showing the experimental data, the fitted Cheeseman–Bennett model, and the fitted proposed model for a nozzle angle of $45^{\circ}$, with fit parameters for different nozzle angles shown.}
    \label{fig:cheeseman}
    \vspace{-2em}
\end{figure}

To assess the applicability of classical ground-effect models to a shrouded multirotor configuration, the experimental data were fitted to the Cheeseman–Bennett model \cite{cheeseman_effect_1955}, as shown in Fig.~\ref{fig:cheeseman}, with the governing equation given as: 

\begin{equation}
\frac{F_{\mathrm{IGE}}}{F_{\mathrm{OGE}}} = \frac{1}{1 - (\frac{k}{Z/r})^2}
\end{equation}

Curve fitting was performed independently for nozzle angles of $15^\circ$, $30^\circ$, $45^\circ$, and $60^\circ$. Despite substantial variation in the measured force–height profiles across nozzle angles, the fitted parameter $k$ converged to nearly identical values in all cases. This insensitivity to geometric variation indicates that the model lacks sufficient expressive power to capture the geometry-dependent aerodynamic interactions introduced by the shroud. In particular, the Cheeseman–Bennett formulation does not reproduce the pronounced suckdown dip observed during transition from in-ground-effect (IGE) to out-of-ground-effect (OGE) operation. This limitation is consistent with prior studies on small-scale rotorcraft, which report deviations from image-theory-based ground effect models when multi-rotor coupling, flow separation, and low-Reynolds-number effects become significant~\cite{11}. More advanced modeling approaches incorporating geometry-dependent pressure fields and unsteady flow effects are required for a comprehensive physical description. As an initial step, we propose an empirically derived model that captures both in-ground-effect thrust enhancement and suckdown behavior:
\begin{equation}
\frac{F_{\mathrm{IGE}}}{F_{\mathrm{OGE}}} = 1 + \frac{A}{1 + \frac{Z/r}{k}} - B e^{-Z/r},
\end{equation}
where $A$ is correlated with the ground-effect thrust enhancement and $B$ with the suckdown contribution; $k$ controls the decay rate with normalized height.

Extracted fit parameters for Equation 2 are shown for different nozzle angles in Figure~\ref{fig:cheeseman}. They provide a quantitative basis for rapid comparison of different designs; for example, it is evident that suckdown magnitude (correlated with $B$) increases monotonically with nozzle angle, while in-ground-effect force amplification does not benefit from angle increasing beyond $45^\circ$. Largely similar $k$ values indicate near equivalence in out-of-ground-effect performance.

\subsection{Hoverflie Dual-Mobility and Tracking Error}

\begin{figure*} 
    \centering
    \includegraphics[width=1\linewidth]{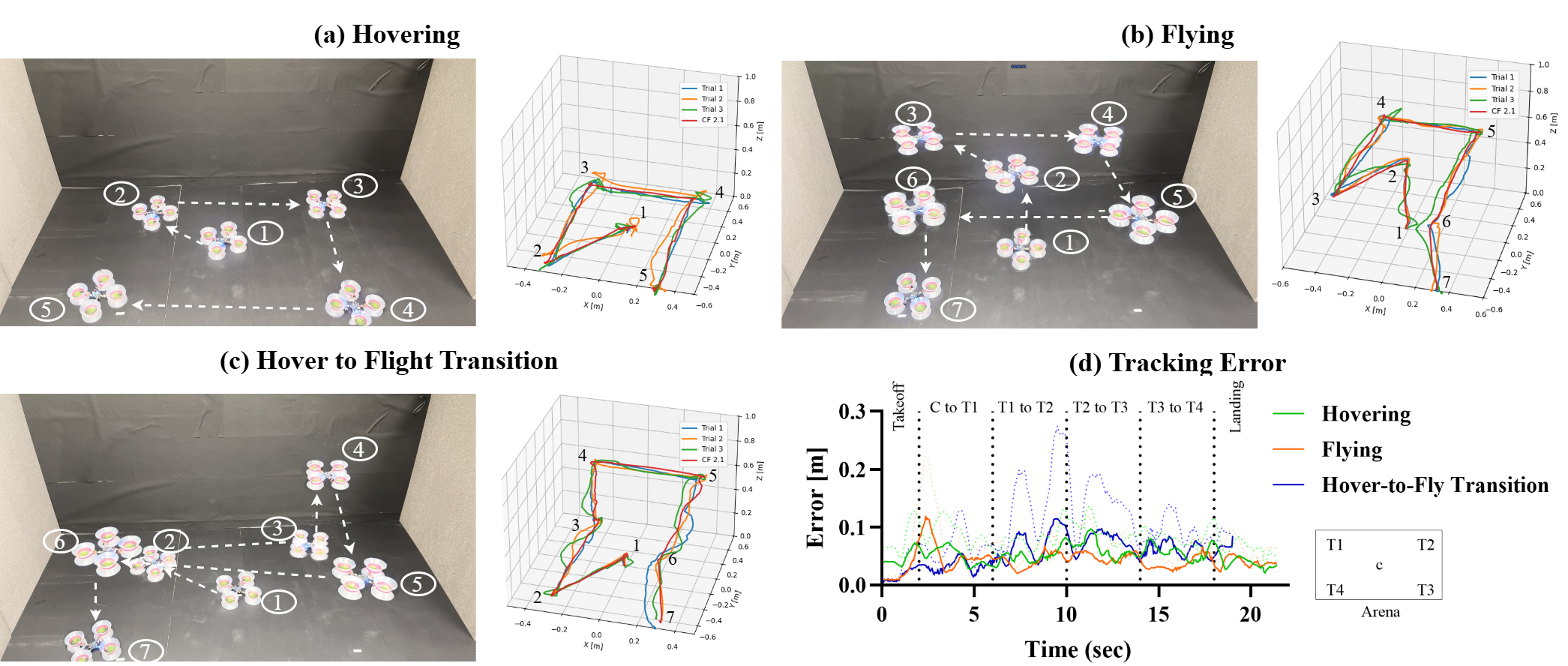}
    \caption{Time-series overlays of Hoverflie during (a) hovering, (b) flying, (c) hover-to-flight transitions, and (d) tracking error during hovering, flying, and hover-to-fly transition maneuvers. Solid lines show mean error over three trials; dotted lines indicate positive standard deviation.}
    \vspace{-1.5em}
    \label{fig:trajectories}
\end{figure*}

The Hoverflie were controlled using a custom Python interface with global $x$-$y$-$z$ feedback provided by a Lighthouse positioning system. Three representative maneuvers were experimentally validated to demonstrate hybrid mobility and coordinated operation (Fig.~\ref{fig:trajectories}). For the controlled flight experiments, the Crazyflie 2.1 Brushless model is used, as the Brushed model struggles to fly stably out-of-ground-effect with the added mass of the shroud. 

In the near-ground hovering maneuver (Fig.~\ref{fig:trajectories}(a)), the vehicle maintains a nearly constant altitude while tracking a planar trajectory. This demonstrates the intrinsic stability of proximity-effect vehicles; as the motor duty cycle required for hovering is substantially lower than the free-flight value, it passively returns to an equilibrium hover height without significant control effort required for maintaining altitude.

In free-flight (Fig.~\ref{fig:trajectories}(b)), the vehicle tracks a three-dimensional trajectory at an altitude out-of-ground-effect. Increased thrust authority and active altitude compensation are required to maintain trajectory stability outside the ground effect region. Yaw control authority was noticeably impacted by the addition of the shrouds; whether this was solely from the increased inertia or due in part to more complicated fluid mechanic effects remains an area of future study. 

The hover-to-flight transition (Fig.~\ref{fig:trajectories}(c)) demonstrates smooth regulation across the ground-effect boundary. To achieve this, the attitude PID controller parameters onboard the Hoverflie are adjusted once the vehicle ascends out-of-ground-effect. This switched mode control architecture is limited in that it requires tuning for strongly in-ground-effect and fully out-of-ground-effect operation, and requires avoiding flight in the transition region entirely. 

The tracking performance of the Hoverflie was evaluated by computing the Euclidean error relative to stock Crazyflie 2.1 reference trajectories during hovering, flying, and hover-to-fly transition maneuvers (Fig.~\ref{fig:trajectories}(d)). The graph shows that the hovering maneuver maintained relatively low tracking error comparable to the reference, while the flying maneuver exhibited larger errors primarily during takeoff, indicating the need for more robust control strategies during this transition phase. The hover-to-fly transition showed the highest deviation due to rapid aerodynamic changes between in-ground-effect and out-of-ground-effect operation. The RMS error during hovering, flying, and hover-to-fly transitions was 6.0 cm, 5.3 cm, and 7.8 cm, respectively. 

\subsection{Structural Mass Changes}

The conversion from Crazyflie to Hoverflie required structural modifications that introduced additional components while removing the original propeller guards. As summarized in Table I, the added structural elements contribute 21.2~g and 5.4~g is reduced from removed components, resulting in a net mass increase of 15.8~g, approximately 50\% of the Crazyflie body weight. This breakdown illustrates the value of thermoforming, as the large volume shrouds are only a relatively small fraction of the total mass. It also shows the opportunity to further improve performance using lighter arms and extenders (e.g., with carbon fiber).

\begin{table}
\caption{Mass Modification Summary for Hoverflie Conversion}
\label{tab:mass_budget}
\centering
\footnotesize
\begin{tabular}{l c c}
\hline
Component & Calc. (g) & $\Delta$ Mass (g) \\
\hline
3D-printed arms & $4\times3.1$ & +12.4 \\
Thermoformed shroud & $4\times1.2$ & +4.8 \\
Extension connectors & $4\times1.0$ & +4.0 \\
\textbf{Added subtotal} &  & \textbf{+21.2} \\
Propeller guards (removed) & $4\times1.34$ & -5.4 \\
\textbf{Net mass increase} &  & \textbf{+15.8} \\
\hline
\end{tabular}
\vspace{-2em}
\end{table}

\subsection{Comparative Flight Endurance Analysis}

\begin{figure}
    \centering
    \includegraphics[width=1\linewidth]{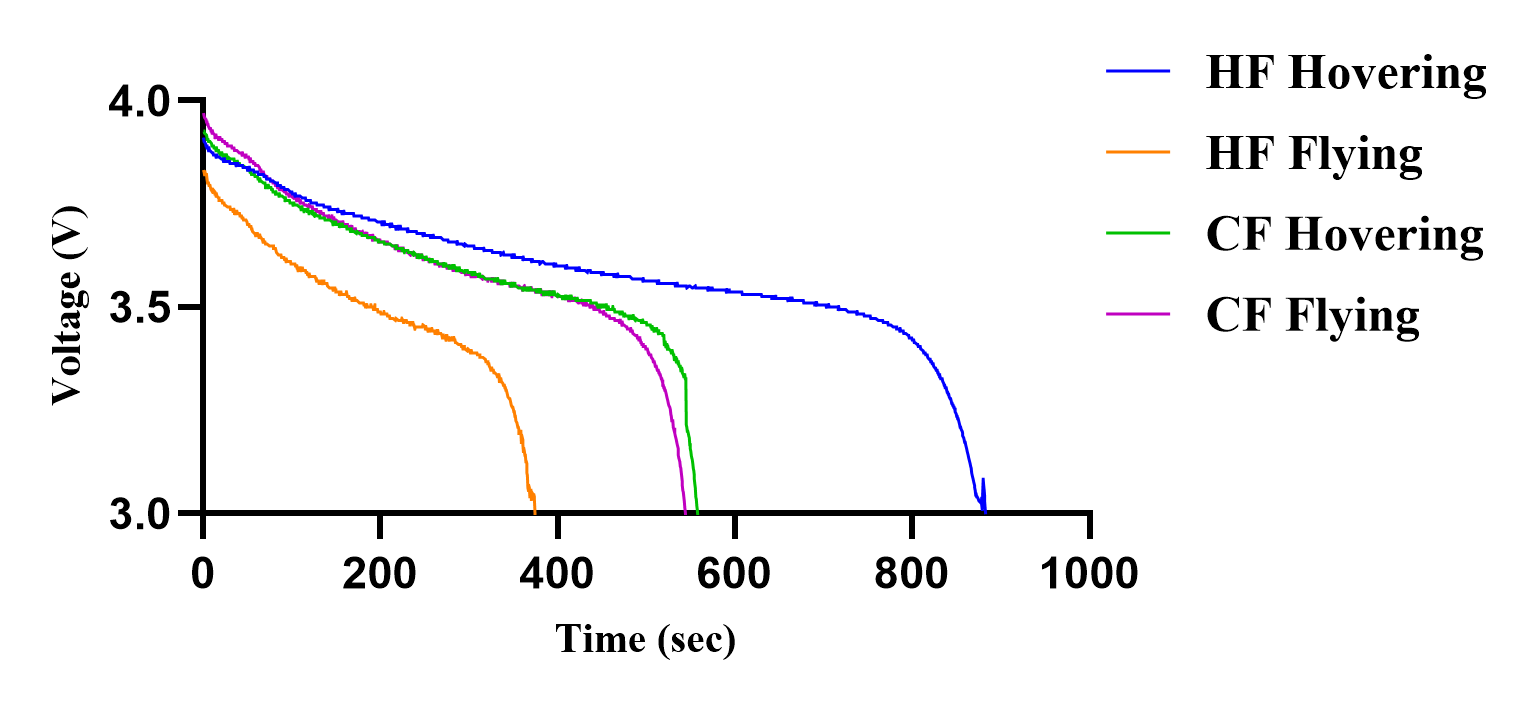}
    \vspace{-3em}
    \caption{Battery voltage decay and flight duration for Hoverflie hovering, Hoverflie flying, Crazyflie hovering, and Crazyflie flying.}
    \label{fig:flighttime}
    \vspace{-2em}
\end{figure}

Battery voltage decay profiles for four configurations are shown in Fig.~\ref{fig:flighttime}: Hoverflie hovering in-ground-effect (HF Hovering), Hoverflie in free-flight (HF Flying), stock Crazyflie 2.1 Brushless hovering near the ground (CF Hovering), and stock Crazyflie in free-flight (CF Flying). 

In-ground-effect hovering, Hoverflie achieved a flight duration of 889 s, compared to 561 s for the Crazyflie hovering near the ground. In contrast, during free-flight, Hoverflie operated for 382 s, whereas the stock Crazyflie achieved 544 s. These results correspond to an approximately 60\% improvement in endurance during in-ground-effect operation, accompanied by a 30\% reduction in free-flight endurance. This quantifies the central trade-off of the hybrid architecture: substantial energy savings in proximity-assisted operation with a penalty during full aerial flight. 



\section{Limitations and Future Work}
\label{sec:Limitations and Future Work}

This study isolates vertical force generation using a quasi-static, single-axis configuration. Consequently, the characterization of thrust versus height from the ground plane does not capture unsteady effects from lateral motion, attitude variation, or aggressive hover-to-flight transitions. In addition, experiments were conducted above a rigid, flat surface in controlled indoor conditions. Surface compliance and texture or environmental disturbances (e.g., cross-flow) may alter pressure distributions, for example. 

The shroud geometries explored represent a structured but limited subset of the full design space. Although duct diameter, height, nozzle angle, and intake configuration were varied systematically, no full multifactorial optimization was performed, and interactions between parameters may extend beyond those captured here.

Future work will extend this initial framework to dynamic ground-effect characterization during horizontal motion and attitude variation, along with broader exploration of the coupled geometric design space, enabling integration into a more robust control architecture.

\section{Conclusions}
\label{sec:conclusions}

This paper presented an empirical investigation of shroud geometry for transforming a micro air vehicle into a ground-effect–dominant hybrid hovercraft. Using controlled vertical force characterization on a reproducible Crazyflie-based platform, we identified how duct confinement, nozzle shaping, and intake geometry govern lift amplification, suction-induced penalties, and free-flight performance.

A key finding is the tradeoff between lift enhancement and transition stability, where geometric modifications that increase in-ground-effect lift can also amplify adverse suckdown forces. By balancing these effects, an optimized configuration achieved nearly three times higher normalized lift in-ground-effect without a large degradation in out-of-ground-effect thrust production. Classical helicopter-derived ground-effect models were shown to be insufficient for capturing the nonlinear behavior of shrouded multirotor systems; we propose an empirical model that captures the resulting suckdown effect, and show it allows for simple comparison between different design configurations. Ground-effect operation increased Hoverflie endurance by approximately 60\% compared to stock Crazyflie flight and multi-modal locomotion (i.e., efficient near-surface hovering and controlled free-flight) is demonstrated with a simple mode-switching control architecture. 

Overall, this work establishes passive structural modification as a practical design lever for energy-aware mobility in micro aerial systems, and introduces an extendable platform for future experimentation. 




\bibliographystyle{IEEEtran}

\bibliography{hoverflie}

\end{document}